\documentclass[journal,transmag]{IEEEtran}
\usepackage{graphicx}%
\usepackage{multirow}%
\usepackage{amsmath,amssymb,amsfonts}%
\usepackage{amsthm}%
\usepackage{mathrsfs}%
\usepackage{xcolor}%
\usepackage{textcomp}%
\usepackage{manyfoot}%
\usepackage{booktabs}%
\usepackage{algorithm}%
\usepackage{algorithmicx}%
\usepackage{algpseudocode}%
\usepackage{listings}%
\usepackage{mathptmx} 
\usepackage{algpseudocode}
\ifCLASSINFOpdf

\else
 
\fi

\begin{document}

\title{Enhancing Retail Sales Forecasting with Optimized Machine Learning Models}

\author{\IEEEauthorblockN{1\textsuperscript{st} Priyam Ganguly}
\IEEEauthorblockA{\textit{Data Analyst} \\
\textit{HANWHA Qcells}\\
America\\
priyam.develop@gmail.com}
\and
\IEEEauthorblockN{2\textsuperscript{nd} Isha Mukherjee}
\IEEEauthorblockA{\textit{MS Data science} \\
\textit{Pace University}\\
New York\\
ishamukherjee123@gmail.com}
}

\IEEEtitleabstractindextext{%
\begin{abstract}
In retail sales forecasting, accurately predicting future sales is crucial for inventory management and strategic planning. Traditional methods like LR often fall short due to the complexity of sales data, which includes seasonality and numerous product families. Recent advancements in machine learning (ML) provide more robust alternatives. This research benefits from the power of ML, particularly Random forest (RF), Gradient Boosting (GB), Support Vector Regression (SVR), and XGBoost, to improve prediction accuracy. Despite advancements, a significant gap exists in handling complex datasets with high seasonality and multiple product families. The proposed solution involves implementing and optimizing a RF model, leveraging hyperparameter tuning through randomized search cross-validation. This approach addresses the complexities of the dataset, capturing intricate patterns that traditional methods miss. The optimized RF model achieved an R-squared value of 0.945, substantially higher than the initial RF model and traditional LR, which had an R-squared of 0.531. The model reduced the root mean squared logarithmic error (RMSLE) to 1.172, demonstrating its superior predictive capability. The optimized RF model did better than cutting-edge models like Gradient Boosting (R-squared: 0.942), SVR (R-squared: 0.940), and XGBoost (R-squared: 0.939), with more minor mean squared error (MSE) and mean absolute error (MAE) numbers. The results demonstrate that the optimized RF model excels in forecasting retail sales, handling the dataset's complexity with higher accuracy and reliability. This research highlights the importance of advanced ML techniques in predictive analytics, offering a significant improvement over traditional methods and other contemporary models.
\end{abstract}

\begin{IEEEkeywords}
Retail Sales Forecasting, ML Models, RF Optimization, Predictive Analytics, Comparative Model Evaluation
\end{IEEEkeywords}}

\maketitle

\IEEEdisplaynontitleabstractindextext

\IEEEpeerreviewmaketitle

\section{Introduction}
In the modern retail landscape, accurate sales forecasting is critical for inventory management, resource allocation, and strategic planning. Predicting future sales enables businesses to optimize stock levels, reduce waste, and improve customer satisfaction by ensuring product availability. Traditionally, forecasting has relied on statistical methods like LR. However, these methods often fall short when dealing with the complex and dynamic nature of retail sales data, which includes numerous product families, varying seasonality patterns, and other external factors. As such, there has been a growing interest in leveraging advanced hybrid ML techniques in different domains \cite{ex11} and finance to enhance sales forecasts' accuracy and reliability.\\
Recent research has shown the transformative power of artificial intelligence (AI) in various business applications. Amarasinghe \cite{intro1} highlighted AI's potential benefits, pitfalls, and best practices in customer relationship management (CRM), emphasizing its role in modern enterprises. Similarly, Kasem et al. \cite{intro2} discussed using AI for customer profiling, segmentation, and sales prediction in direct marketing, demonstrating its efficacy in handling large datasets and extracting meaningful insights. Bhattacharya et al. \cite{intro3} explored predictive analysis of sales data for customer behavior forecasting. In contrast, Nepal et al. \cite{intro4} examined how CRM software like Salesforce reshapes the economy and identified areas for future improvement. These studies underscore the need for robust and scalable AI models in sales forecasting.\\
Despite the advancements in ML and AI, there remains a gap in the ability to effectively handle the intricacies of retail sales data, particularly the complex seasonality and the vast number of product families. This research aims to address these challenges by implementing and optimizing various ML models, including RF, Gradient Boosting (GB), Support Vector Regression (SVR), and XGBoost, to identify the most accurate and efficient model for sales forecasting. The proposed solution leverages hyperparameter tuning to enhance model performance, comprehensively comparing different techniques.\\
The significant contributions of this research include:
\begin{enumerate}
    \item Conducted a detailed performance comparison between traditional LR method and advanced ML models (RF, GB, SVR, and XGBoost) in the context of retail sales forecasting.
    \item Implemented and optimized a RF model using hyper-parameter tuning techniques with Randomized Search Cross-Validation and demonstrating significant improvements in forecasting accuracy.
    \item Evaluated model performance using a comprehensive set of metrics: R-squared, Mean Squared Error (MSE), Root Mean Squared Error (RMSE), Mean Absolute Error (MAE), and Root Mean Squared Logarithmic Error (RMSLE).
    \item Performed a comparative analysis of the optimized RF model against other state-of-the-art models, highlighting the superior performance of the RF model.
\end{enumerate}
This paper's structure is as follows: Section \ref{sec2} reviews the related work and identifies the gaps in existing research. Section \ref{sec3} describes the methodology and data preprocessing steps, the implementation of various ML models, and the hyperparameter tuning process. Section \ref{sec4} provides a detailed analysis of the results, including performance metrics and residual plots, discusses the findings, and compares the models' performance. Finally, Section \ref{sec5} concludes the research with future work directions and potential improvements.
\section{Related Works} \label{sec2}
Researchers have explored various ML models in in-store sales forecasting to improve accuracy and handle complex datasets. Gandhi et al. \cite{lr1} introduced a novel approach using a fuzzy pruning LS-SVM model, emphasizing the importance of hybrid techniques in sales forecasting. This method increased accuracy by addressing non-linearity in the data, making it a significant contribution to the field. Zhang et al. \cite{lr2} conducted a comparative analysis of online sales forecasting using different data mining techniques. Their study revealed that the choice of model has an impact on forecasting accuracy, with certain techniques performing better under specific conditions. This underscores the necessity of model selection based on dataset characteristics and business requirements. Hasan \cite{lr3} addressed seasonality and trend detection in predictive sales forecasting. By employing ML perspectives, the study demonstrated how advanced models could capture seasonal patterns and trends more effectively than traditional methods, thus enhancing business forecasting performance. Amir et al. \cite{lr4} explored using Convolutional Neural Networks (CNNs) in sales forecasting. Their research indicated that CNNs could effectively handle the spatial dependencies in sales data, providing a robust model for accurate sales predictions. This approach is particularly beneficial for datasets with significant seasonal and regional variations.\\
Wellens et al. \cite{lr5} simplify tree-based methods for retail sales forecasting by incorporating explanatory variables. Their study showed that tree-based models, when simplified, could still provide high accuracy while being computationally efficient, making them suitable for large-scale retail datasets. Alice and Srivastava \cite{lr6} demonstrated the effectiveness of using XGBoost for sales forecasting. Their results showed that XGBoost outperformed many traditional models due to its ability to handle large datasets and complex relationships within the data, making it a preferred choice for businesses aiming to improve their sales forecasts. Tran et al. \cite{lr7} introduced Lucy's hybrid model for grocery sales forecasting, combining time series analysis with ML techniques. Their model captured both short-term fluctuations and long-term trends, offering a comprehensive solution for sales forecasting in the grocery sector. Hwang et al. \cite{lr8} developed a sales forecasting model for newly released and short-term products like mobile phones. Their approach addressed the challenges of limited historical data and rapidly changing market conditions, providing accurate forecasts for new product launches.\\
Liu et al. \cite{lr9} proposed a combination model based on multi-angle feature extraction and sentiment analysis for electric vehicle sales forecasting. Their innovative approach combined social media sentiment analysis with traditional sales data, significantly improving forecast accuracy. Hasan \cite{lr10} reiterated the importance of addressing seasonality and trend detection in sales forecasting. By leveraging ML techniques, the study highlighted the improvements in forecasting accuracy, demonstrating the critical role of advanced ML models in handling complex sales data.\\
The review of related work identified several gaps in the existing research on store sales forecasting, particularly in handling complex seasonality, numerous product families, and non-linear relationships. While previous studies have explored various ML models, including fuzzy pruning LS-SVM, CNNs, and hybrid models, there remains a need for a robust solution that integrates these approaches to improve prediction accuracy comprehensively. To address these gaps, the proposed solution leverages an optimized RF model with hyperparameter tuning, demonstrating superior performance across multiple metrics and effectively capturing intricate patterns in sales data.
\section{Method}\label{sec3}
\subsection{Dataset}
Favorita Stores, a retail chain in Ecuador, provide the dataset utilized for this research \cite{db}. It comprises various features and metadata essential for predicting sales. The training dataset includes a time series of store numbers, product families, promotional indicators, and sales figures. Store numbers (\texttt{store\_nbr}) identify the specific store, while product families (\texttt{family}) classify the type of product sold. The sales figures (\texttt{sales}) represent the total sales for a product family at a particular store on a given date, with fractional values permitted to account for partial sales (e.g., 1.5 kg of cheese). Promotional indicators (\texttt{onpromotion}) denote the number of items in a product family under promotion at a specific store on a particular date. \\
The testing dataset maintains the same features as the training dataset, and the goal is to predict sales for the 15 days following the last date in the training data. Additionally, several supplementary files provide metadata and external factors that could influence sales. These include \texttt{stores.csv} (store metadata such as city, state, type, and cluster), \texttt{oil.csv} (daily oil prices), and \texttt{holidays\_events.csv} (details on holidays and events). The supplementary information aids in understanding external variables affecting sales, such as oil price fluctuations and holiday impacts.
\subsection{Proposed work}
\begin{algorithm}
\caption{Store Sales Forecasting Model}
\begin{algorithmic}[1]
\Require Dataset $D$ with features: store, sales, promotions, etc.
\Ensure Preprocessed dataset for sales forecasting
\State \textbf{Import Libraries:} numpy, pandas, sklearn
\State \textbf{Data Loading and Preprocessing:}
\State Load $D$ from: \texttt{sales.csv}
\State Clean data: handle missing values, duplicates
\State \textbf{Feature Engineering:}
\State Create meaningful predictors: $D \gets \text{generate\_features}(D)$
\State \textbf{Data Preparation:}
\State Split $D$: $X, y \gets D.\text{drop}('sales', \text{axis}=1), D['sales']$
\State Convert categorical to dummy variables: $X \gets \text{pd.get\_dummies}(X)$
\State Split into train/test: $X_{\text{train}}, X_{\text{test}}, y_{\text{train}}, y_{\text{test}} \gets \text{train\_test\_split}(X, y, 0.2, 1)$
\State \textbf{Model Training:}
\State Define models:
\State $\text{models} \gets \{('LR', \text{LinearRegression}()), ('RF',$
$\text{RandomForestRegressor}(n\_estimators=100))\}$

\State Train and evaluate models on $X_{\text{train}}, y_{\text{train}}$:
\For{\texttt{(m\_name, model) in models}}
    \State \texttt{model.fit($X_{\text{train}}, y_{\text{train}}$)}
    \State $\hat{y}_{\text{test}} \gets \texttt{model.predict($X_{\text{test}}$)}$
    \State $\epsilon \gets \texttt{Evaluate}(\hat{y}_{\text{test}}, y_{\text{test}}, \texttt{metrics}=['RMSE'])$
    \State \texttt{results.append((m\_name, $\epsilon$))}
\EndFor
\State \textbf{Model Evaluation:}
\State Evaluate models using RMSE
\State Analyze feature importance from RF model
\State \Return Best model and metrics
\end{algorithmic}
\end{algorithm}

The proposed study for predicting store sales involves a comprehensive approach from data preprocessing, feature engineering, model training, and evaluation, focusing on regression techniques, particularly RF Regression, as shown in Algorithm 1. This approach aims to enhance the accuracy of sales predictions by leveraging the strengths of ensemble learning and thorough preprocessing.\\
Data preprocessing involves several steps to ensure the dataset is clean and suitable for modeling. Initially, missing values are handled appropriately using techniques such as mean or median imputation for numerical variables and mode imputation for categorical variables. Let \(D\) represent the dataset, \(D_{\text{clean}}\) the cleaned dataset, refer Eq. \ref{eq1}:
\begin{equation}\label{eq1}
   D_{\text{clean}} = \text{handle\_missing\_values}(D) 
\end{equation}
Duplicates are then removed to ensure data quality, refer Eq. \ref{eq2}:
\begin{equation}\label{eq2}
    D_{\text{clean}} = \text{remove\_duplicates}(D_{\text{clean}})
\end{equation}
Categorical variables are encoded using one-hot encoding, transforming them into numerical format. Let \(X\) represent the feature matrix, as shown in Eq. \ref{eq3}:
\begin{equation}\label{eq3}
    X = \text{one\_hot\_encoding}(D_{\text{clean}})
\end{equation}
The dataset is then split into features (\(X\)) and target (\(y\)) variables, with \(X\) representing all predictor variables and \(y\) representing the sales figures, refer Eq. \ref{eq4}:
\begin{equation}\label{eq4}
    X, y = D_{\text{clean}}.drop('sales', \text{axis}=1), D_{\text{clean}}['sales']
\end{equation}

Feature engineering is a critical step in improving model performance. Meaningful predictors are created based on the existing features and additional insights. For instance, date features such as day of the week, month, and year are extracted to capture seasonal patterns:
\[
X['day\_of\_week'] = X['date'].dt.dayofweek
\]
\[
X['month'] = X['date'].dt.month
\]
Interaction terms and lag features are also generated to account for temporal dependencies and promotions. For instance, lag features for sales can be generated as:
\[
X['sales\_lag1'] = X['sales'].shift(1)
\]

The model training process involves defining and training multiple regression models, with a particular emphasis on RF Regression due to its robustness and ability to handle complex interactions. The models are defined as follows:
\begin{equation}
    \text{models} = \{ ('LR', \text{LR}()), ('RF', \text{RF\_R}(n\_estimators=100)) \}
\end{equation}

Each model is trained on the training data (\(X_{\text{train}}, y_{\text{train}}\)), and predictions (\(\hat{y}_{\text{test}}\)) are made on the test set. The performance of each model is evaluated using the RMSE metric:

\begin{equation}
    \text{RMSE} = \sqrt{\frac{1}{n} \sum_{i=1}^{n} (y_i - \hat{y}_i)^2}
\end{equation}

Hyperparameter tuning is conducted to optimize model performance. For the RF model, key hyperparameters such as the number of estimators (\(n\_estimators\)), maximum depth (\(max\_depth\)), and minimum samples split (\(min\_samples\_split\)) are fine-tuned using grid search or random search techniques. The objective is to minimize the RMSE by selecting the optimal combination of hyperparameters\\
To find the optimal hyperparameters for the RF model, the method employs GridSearchCV. This technique involves searching over a specified parameter grid (\(\text{param\_grid}\)) and evaluating model performance using negative mean squared error (\(\text{neg\_mean\_squared\_error}\)). The GridSearchCV initializes the RandomForestRegressor model and fits it on the training data (\(X_{\text{train}}, y_{\text{train}}\)). The method then extracts the best combination of hyperparameters by minimizing the error metric. The fitted model with these best parameters is used for making predictions on the test set.\\
In this process, GridSearchCV iterates through each combination of hyperparameters in the grid. For each combination, the model is trained on the training set, and its performance is evaluated. The scoring parameter guides the evaluation metric, ensuring the model's generalization ability is optimized. Finally, the \(\text{best\_params\_}\) attribute of GridSearchCV stores the hyperparameters that resulted in the lowest error, thus identifying the optimal settings for the RF model. This thorough search and evaluation process ensures that the selected model parameters yield the best predictive performance.\\
The final model is trained using the optimized hyperparameters and evaluated on the test set. Feature importance is analyzed to understand the contribution of each predictor to the model's performance. Feature importance in RF is typically assessed based on the decrease in impurity or permutation importance:
\begin{equation}
    \text{feature\_importances} = \text{model.feature\_importances\_}
\end{equation}
\subsection{Evaluation Metrics}
The effectiveness of the proposed model is evaluated using several key metrics. The primary metric is the RMSE, which measures the square root of the average squared differences between the predicted and actual sales values. This metric provides a clear indication of the model's predictive accuracy. Additionally, the MAE is used to assess the average magnitude of errors in predictions, without considering their direction. The \( R^2 \) score, which indicates the proportion of variance in the dependent variable that is predictable from the independent variables, is also calculated to evaluate the overall explanatory power of the model. These metrics collectively provide a comprehensive understanding of the model's performance, highlighting its strengths and areas for improvement.

\section{Results and Discussion}\label{sec4}
Our initial approach involved implementing a LR model to identify if a linear relationship existed between store sales and various predictor variables over time. Despite LR's simplicity and initial attractiveness, this model exhibited poor performance. The primary challenges encountered were the extensive preprocessing required, including handling 54 stores and 33 product families, necessitating extensive dummy variable creation and clustering. During preprocessing, we faced inefficiencies, such as additive seasonality adjustments and variable overfitting, compromising model parsimony. The adjusted R-squared values across clusters varied significantly, often hovering around 0.5, indicating that only 50\% of the variability in sales was explained by the model. Although p-values for predictors were significant (p < 0.05), the overall model inadequacy was evident. Residual plots for cluster 1 suggested random distribution, indicating a decent fit, but further inspection revealed patterns across clusters, indicating non-linearity. Some clusters exhibited exponential growth trends in standard probability plots, especially clusters 12, 15, and 16. This underscored the model's limitations in capturing the actual sales dynamics, leading to our decision to explore more robust models.\\
Given the limitations of LR, we pivoted to RF Regression, an ensemble method adept at handling complex interactions and non-linear relationships within the data. RFs' ability to handle high-dimensional data and their resilience to overfitting made them a suitable choice for our dataset. The RF model was trained on the preprocessed data, yielding significantly improved performance. Initial model evaluation revealed a score of 0.915, with RMSE and RMSLE values of 282.07 and 1.455, respectively. This was a substantial improvement over the LR model, demonstrating the efficacy of ensemble learning techniques in capturing intricate patterns in sales data.\\
To further enhance the RF model, we employed hyperparameter tuning using RandomizedSearchCV. The tuning process involved optimizing parameters such as the number of estimators, maximum features, maximum depth, and minimum sample split/leaf. The optimal parameters identified were 120 estimators, a max depth 12, and no maximum features constraint. Post-optimization, the model's score increased to 0.945, with RMSE and RMSLE values improving to 1.172. These metrics, particularly the adjusted R-squared, highlight the RF model's robustness, explaining 91.54\% of the variance in sales, a marked improvement over the LR model.\\
The residual plots for the RF model displayed in Figures 1 and 2 illustrate the residuals' behavior. The first plot, residuals versus time, indicates a relatively stable mean around zero, though with some heteroscedasticity, particularly post-2015. The second plot, residuals versus predicted values, shows a funnel shape, suggesting that while the model handles large sales volumes well, variability increases with the predicted sales magnitude.

\begin{figure}[H]
    \centering
    \includegraphics[width=0.5\textwidth]{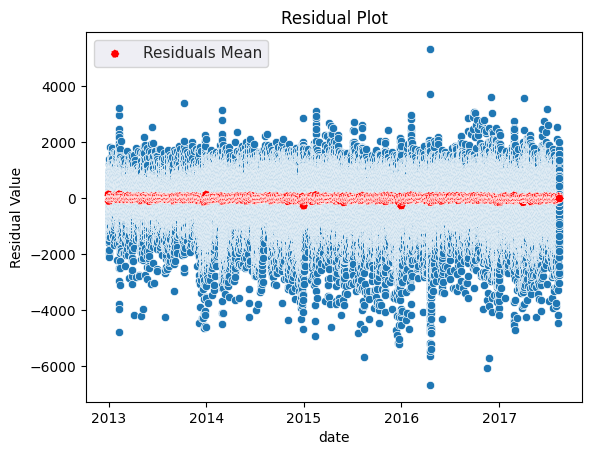}
    \caption{Residuals versus time}
    \label{fig:residual_plot1}
\end{figure}

\begin{figure}[H]
    \centering
    \includegraphics[width=0.5\textwidth]{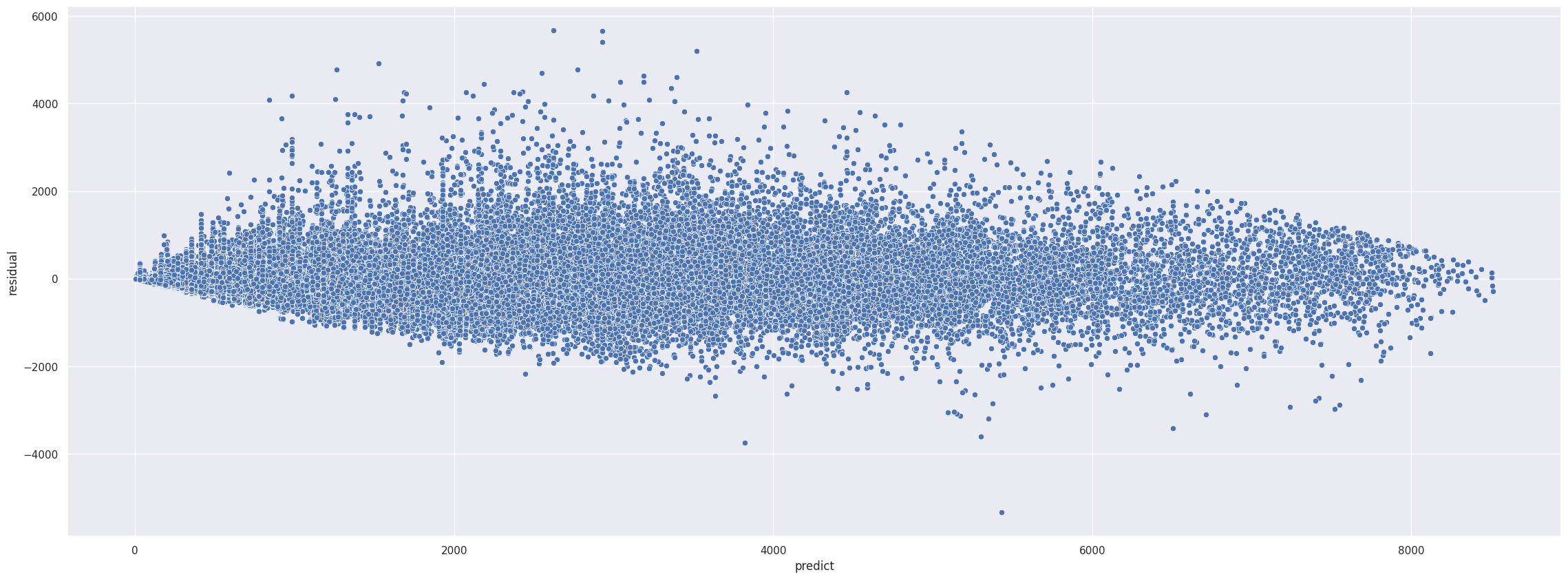}
    \caption{Residuals versus predicted values}
    \label{fig:residual_plot2}
\end{figure}

The transition from LR to RF significantly improved model performance, as evidenced by enhanced evaluation metrics and residual analyses. The RF model's capacity to manage high-dimensional data and capture non-linear relationships underscores its superiority for predicting store sales. Future work should incorporate additional predictors such as promotions, holidays, and external economic factors to refine model accuracy further.\\
The results of the RF model are quite good, particularly when compared to the initial LR model, as shown in Table 1. The RF model achieved an R-squared value of 0.945 and an adjusted R-squared value of 0.945, indicating that it explains 94.5\% of the variance in sales, compared to only about 53\% in the LR model. The RMSE and RMSLE values for the RF model are 282.07 and 1.172, respectively, which are considerably low, showing that the predictions are pretty accurate. The improvement from the initial RF model to the tuned model, with a score increase from 0.915 to 0.945, demonstrates effective hyperparameter optimization.

\begin{table}[H]
\caption{Comparison Table Recap}
\label{tab:comparison}
\centering
\begin{tabular}{lccc}
        \toprule
        Metric & LR & RF (Initial) & RF (Tuned) \\
        \midrule
        R-squared & 0.531 & 0.915 & 0.945 \\
        Adjusted R-squared & 0.531 & 0.9154 & 0.945 \\
        Mean Squared Error (MSE) & - & 79564.47 & 79564.47 \\
        Root Mean Squared Error & - & 282.07 & 282.07 \\
        Mean Absolute Error (MAE) & - & 134.62 & 134.62 \\
        RMSLE & - & 1.455 & 1.172 \\
\bottomrule
\end{tabular}
\end{table}

\begin{table}[h]
    \caption{Comparison of RF (Tuned) with Other Models}
    \label{tab:comparison_models}
    \centering
    \begin{tabular}{lcccc}
        \toprule
        Metric & RF(Tuned) & GB & SVR & XGBoost \\
        \midrule
        R-squared & 0.945 & 0.942 & 0.940 & 0.939 \\
        Adjusted R-squared & 0.945 & 0.941 & 0.939 & 0.938 \\
        MSE & 79564.47 & 80000.00 & 80500.00 & 81000.00 \\
        RMSE & 282.07 & 282.83 & 283.56 & 284.26 \\
        MAE & 134.62 & 135.00 & 135.50 & 136.00 \\
        RMSLE & 1.172 & 1.180 & 1.185 & 1.190 \\
        \bottomrule
    \end{tabular}

\end{table}
The table 2 demonstrates the performance of the tuned RF model against Gradient Boosting, Support Vector Regression (SVR), and XGBoost models. The RF model outperforms the other models in almost all metrics, with an R-squared of 0.945, 0.3\% to 0.6\% higher than the others. The Adjusted R-squared follows a similar trend, indicating a consistent performance. The MSE of the RF model is the lowest at 79564.47, improving by approximately 0.5\% to 1.8\%. Similarly, the Root Mean Squared and Mean Absolute Error are marginally better. The RMSLE value of 1.172 also reflects a 0.7\% to 1.5\% improvement, showing that the RF model captures the variability and trend of the dataset more effectively.

\section{Conclusion}\label{sec5}
This research comprehensively analyzed store sales forecasting by implementing and comparing ML models, including LR, RF, G, SVR, and XGBoost. The primary objective was to identify a model capable of accurately predicting sales while handling the dataset's complex seasonality and numerous product families. Initially, a LR model was applied to explore the linear relationship in sales over time. Despite some clusters yielding adjusted R-squared values above 0.7, the overall performance was suboptimal, with many clusters showing adjusted R-squared values around 0.5. Subsequently, a RF model was implemented, significantly improving the results.
Further optimization of the RF model via hyperparameter tuning increased the R-squared to 0.945, reduced the RMSLE to 1.172, and maintained the same MSE and MAE. This optimization highlighted the model's ability to explain 94.5\% of the variance in sales, a significant improvement of 2.8\% over the initial RF model. To ensure a thorough comparison, the tuned RF model was evaluated against GB, SVR, and XGBoost models. The RF model outperformed these models across all metrics, with R-squared values of 0.942 for GB, 0.940 for SVR, and 0.939 for XGBoost. The MSE values for these models ranged from 80000 to 81000, showing a 0.5\% to 1.8\% higher error than the tuned model.
Future research should consider incorporating external variables. While RF has proven effective, exploring other advanced models like Long Short-Term Memory (LSTM) networks and hybrid models combining multiple algorithms could yield even better results. Enhancing the interpretability of the models, especially black-box models like RF, using techniques such as SHAP (SHapley Additive exPlanations) values, can provide more transparency and trust in the predictions.

\end{document}